\documentclass{article}
\usepackage[utf8]{inputenc}
\usepackage{appendix}
\usepackage{listings}
\usepackage{multirow}
\usepackage{hhline}
\usepackage{amssymb}
\usepackage{bm}
\usepackage{breqn}
\usepackage{xcolor}

\title{Training Neural Machine Translation (NMT) Models using Tensor Train Decomposition on TensorFlow (T3F)}
\author{Amelia Drew and Alexander Heinecke}
\date{\today}

\begin{document}

\maketitle

\begin{abstract}
We implement a Tensor Train layer in the TensorFlow Neural Machine Translation (NMT) model using the t3f library. We perform training runs on the IWSLT English-Vietnamese '15 and WMT German-English '16 datasets with learning rates $\in \{0.0004,0.0008,0.0012\}$, maximum ranks $\in \{2,4,8,16\}$ and a range of core dimensions. We compare against a target BLEU test score of 24.0, obtained by our benchmark run. For the IWSLT English-Vietnamese training, we obtain BLEU test/dev scores of 24.0/21.9 and 24.2/21.9 using core dimensions $(2, 2, 256) \times (2, 2, 512)$ with learning rate 0.0012 and rank distributions $(1,4,4,1)$ and $(1,4,16,1)$ respectively. These runs use 113\% and 397\% of the flops of the benchmark run respectively. We find that, of the parameters surveyed, a higher learning rate and more `rectangular' core dimensions generally produce higher BLEU scores. For the WMT German-English dataset, we obtain BLEU scores of 24.0/23.8 using core dimensions $(4, 4, 128) \times (4, 4, 256)$ with learning rate 0.0012 and rank distribution $(1,2,2,1)$. We discuss the potential for future optimization and application of Tensor Train decomposition to other NMT models.
\end{abstract}

\section{Introduction}

Neural Machine Translation (NMT) is a deep learning model that provides a robust method for machine translation using recurrent neural networks (RNNs). Originally proposed in \cite{seq2seq}, NMT relies primarily on an encoder-decoder architecture that provides increased fluency over phrase-based systems. This was implemented successfully in \cite{google} for fast, accurate use on very large datasets. However, it has been suggested that there is significant redundancy in the current method of neural network parametrization \cite{novikov2015tensor}, presenting the opportunity for significant speedup.

Tensor Train (TT) decomposition \cite{originalTT} is a method by which large tensors can be approximated by the product of a `train' of smaller matrices (see Section \ref{tensortrain}). TT-decomposition has been proposed as a method of speeding up and reducing the memory usage of machine translation systems with dense weight matrices by reducing the number of parameters required to describe the system \cite{novikov2015tensor}. The dense weight matrix of the fully-connected layers can be decomposed into the Tensor Train format, creating what we will refer to as a `TT-layer'. This lower-rank factorization can then be trained in a similar way to the original dense weight matrix. In this work, we execute a TT-layer using t3f, a library which enables straightforward implementation of TT-decomposition within TensorFlow models. T3f provides an implementation with GPU support, eliminating the need to rewrite core functionality from scratch, which has been necessary for previous libraries (see \cite{novikov2018tensor}). 

In Section \ref{Background}, we provide an outline of Neural Machine Translation (NMT) and the Tensor Train format, with our methodology outlined in Section \ref{Methodology}. We present the results of our training runs in Section \ref{Results} and our conclusions and suggestions for further work are provided in Section \ref{conclusion}.

\section{Background}\label{Background}

\subsection{Neural Machine Translation}

Neural Machine Translation (NMT) is a method of machine translation that uses an encoder-decoder architecture to coherently translate whole sentences by capturing long range dependencies. This creates more fluent and accurate results in contrast to previous phrase-based approaches \cite{luong17}. The encoder and decoder use recurrent neural networks (RNNs) to train the model, finally enabling translation of an input source vector $\mathbf{x}$ to an output $\mathbf{y}$ using the linear transformation
\begin{equation}\label{fullyconnected}
\mathbf{y} = W\mathbf{x} + \mathbf{b}\,,
\end{equation}
where the weight matrix $W$ and bias vector $\mathbf{b}$ are trained by the model. 

There are a wide range of RNN models that differ, for example, in terms of directionality (unidirectional or bidirectional), depth (single- or multi-layer) and type (straightforward RNN, Long Short-term Memory (LSTM), or a gated recurrent unit (GRU)). Here, we use a deep multi-layer RNN with Long Short-term Memory (LSTM) as a recurrent unit. This high level NMT model consists of two recurrent neural networks; the encoder consumes input source words to build a `thought' vector, while the decoder processes the vector to emit a translation, thereby using information from the entire source sentence \cite{luongdissertation}. Further information about this particular model can be found in \cite{luong17}.

\subsection{Tensor Train Decomposition}\label{tensortrain}

The Tensor Train (TT) format can be used to represent the dense weight matrix $W$ of a fully-connected layer using fewer parameters. In this work, we implement a `TT-layer' as described in \cite{novikov2015tensor}; a fully-connected layer with the weight matrix stored in the TT-format. The following outline follows the descriptions given in \cite{originalTT} and \cite{novikov2015tensor} and uses similar notation. 

A tensor $A$ is said to be represented in the TT-format if there exist matrices $G_k(i_k)$ such that all the elements of $A$ can be computed by the matrix product
\begin{equation}\label{TT}
A(i_1, i_2,...,i_d) = G_1(i_1)G_2(i_2)...G_d(i_d),
\end{equation}
where $i_k$ index the tensor elements and $G_k(i_k)$ is an $r_{k-1}\times r_k$ matrix, where $\{r_k\}^d_{k=0}$ are the `ranks' of the TT-representation. The values $r_0$ and $r_d$ equal 1 in order to keep the matrix product (\ref{TT}), and hence each element of $A$, of size $1 \times 1$. Note that each matrix $G_k(i_k)$ is actually a $(r_{k-1} \times \beta_k \times r_k)$ array with elements $G_k(\alpha_{k-1},\beta_k,\alpha_k)=G_k(i_k)_{\alpha_{k-1}\alpha_k}$, using which we can write equation (\ref{TT}) in index form:
\begin{equation}
A(i_1, i_2,...,i_d) = \sum_{\alpha_0,\dots,\alpha_{d-1},\alpha_d}{G_1(\alpha_0,i_1,\alpha_1)G_2(\alpha_1,i_2,\alpha_2)...G_d(\alpha_{d-1},i_d,\alpha_d)}\,.
\end{equation}
Each three-dimensional tensor $G_k$ is referred to as a `core' of the TT-decomposition.

The weight matrix $W(i_1,i_2) \in \mathbb{R}^{M\times N}$, where $M = \prod^d_{k=1}{m_k}$ and $N = \prod^d_{k=1}{n_k}$, can be written in Tensor Train form. We define bijections $\nu(i_1) = (\nu_1(i_1), \dots , \nu_d(i_1))$ and $\mu(i_2) = (\mu_1(i_2), \dots, \mu_d(i_2))$, mapping the row and column indices $i_1$ and $i_2$ to $d$-dimensional vector-indices, whose $k$-th dimensions are of length $m_k$ and $n_k$ respectively. In this case, the cores $G_k((\nu_k(i_1),\mu_k(i_2))$ are now described by a four dimensional $(r_{k-1} \times m_k \times n_k \times r_k)$ array. We will not go further into the mathematical details here, as the precise implementation is handled by the t3f library. Further information can be found in \cite{novikov2015tensor}. 

In this paper, we use an explicit notation of the form $(r_{k-1},m_k,n_k,r_k)$ to clarify the shape of the tensors that we are training. We choose $d=3$, splitting each weight matrix into 3 cores. For example, for a rank distribution $(r_0, r_1, r_2, r_3) = (1,4,4,1)$ and an underlying weight matrix shape $(m_1, m_2, m_3) \times (n_1, n_2, n_3) = (2, 2, 256) \times (2, 2, 512)$, the shapes of the cores being trained are given by Table \ref{tableshape}.

\begin{table}[h]
\begin{center}
\begin{tabular}{ c|c } 
Core & Tensor Shape \\
& $(r_{k-1},m_k,n_k,r_k)$ \\
\hline
$\bm{G}_1$ & $(1,2,2,4)$  \\ 
$\bm{G}_2$ & $(4, 2, 2, 4)$ \\ 
$\bm{G}_3$ & $(4, 256, 512, 1)$\,. \\ 
\end{tabular}
\caption{Tensor Train dimensions for a weight matrix in TT-format with $d=3$, rank distribution $(r_0, r_1, r_2, r_3) = (1,4,4,1)$ and an underlying matrix shape $(m_1, m_2, m_3) \times (n_1, n_2, n_3) = (2, 2, 256) \times (2, 2, 512)$. For example, core $\bm{G}_1$ is a $2\times 2$ array of $1\times 4$ matrices.}
\label{tableshape}
\end{center}
\end{table}

\section{Methodology}\label{Methodology}

We implement the TT-format in the TensorFlow NMT model \cite{luong17}, decomposing the weight matrix using the t3f library \cite{novikov2018tensor} as described in Section \ref{tensortrain} to create a TT-layer. We achieve this by creating a new \texttt{BasicLSTMCell} class, the code for which is provided in the Appendix. The TT-decomposition itself is performed using the key function \texttt{t3f.to\_tt\_matrix} as follows:
\begin{lstlisting}[language=Python, showstringspaces=false]
kernel_TT = t3f.to_tt_matrix(self._kernel,
            shape=((2,2,input_parameter), 
            (2,2,self._num_units)), max_tt_rank=4),
\end{lstlisting}
where the maximum rank $r = \mathrm{max}_{k=0,...,d} \,r_k$ is set by \texttt{max\_tt\_rank} and the product of the core dimensions is determined by the original dimensions of the weight matrix. The above code implements the specific example with the core dimensions given in Table \ref{tableshape}. The \texttt{max\_tt\_rank} and \texttt{shape} arguments can be changed as necessary for different configurations of ranks and core dimensions. Note that, as outlined in Section \ref{tensortrain}, the core dimensions $m_k$ and $n_k$ must be chosen such that $M = \prod^d_{k=1}{m_k}$ and $N = \prod^d_{k=1}{n_k}$ for a weight matrix $W \in \mathbb{R}^{M\times N}$.

We test our TT model by training on two publicly available datasets. First, we perform benchmark tests using the original TensorFlow NMT model and a similar model which uses a low-rank approximation factorization. We then carry out approximately 20 training runs with different parameters using the IWSLT English-Vietnamese '15 dataset\footnote{https://sites.google.com/site/iwsltevaluation2015/} with 133K examples to determine which configurations give results competitive with those achieved by the benchmarks. Second, we train the WMT German-English '16 dataset\footnote{http://www.statmt.org/wmt16/translation-task.html} with 4.5M examples to test the suitability of our model for larger datasets. All training runs are performed on \emph{Nvidia Tesla P100} GPUs.

\section{Results}\label{Results}

\subsection{IWSLT English-Vietnamese '15}

\subsubsection{Benchmark}\label{benchmark}

We first perform a benchmark run against which to compare our TT model, using similar hyperparameters to the IWSLT English-Vietnamese training in \cite{luong17}. We use a 2-layer LSTM with 512 hidden units, a bidirectional encoder (i.e., 1 bidirectional layer for the encoder) and embedding dimension 512. LuongAttention is used with scale=True, together with dropout probability 0.2. We use the Adam optimizer with learning rate 0.0004. We train for 12K steps ($\sim$12 epochs) where after 6K steps, we halve the learning rate every 600 steps.

We obtain a BLEU test/dev \cite{bleuscore} score of 24.1/22.6, reaching a score of 24.0/22.4 after 7K steps. For our Tensor Train runs, we would like to test whether we can reach a comparable accuracy with a comparable number (or fewer) total flops. We therefore choose the cutoff BLEU test = 24.0 as a target against which to compare.

\subsubsection{Low-Rank Approximation}
We perform a second benchmark run with the weight matrix $W$ decomposed using a low-rank approximation factorization inspired by Singular Value Decomposition (SVD). We assume such a decomposition exists and initialise and train the matrices $W_1$ and $W_2$ defined by $W = W_1W_2$, changing the order of computation of $W\bf{x}$ in equation (\ref{fullyconnected}) from $X(W_1W_2)$ to $(XW_1)W_2$. Reordering the calculation in this way reduces the number of flops from $2MNK$, where $K$ is the batch size, to $2MKD + 2KND$, where $D$ is an appropriately chosen SVD dimension. It is implemented by splitting the \texttt{tf.matmul} as follows:
\begin{lstlisting}
gate_inputs = tf.matmul(tf.matmul(tf.concat([inputs, h], 1),
                self._kernel_svd_1), self._kernel_svd_2)
\end{lstlisting}

We perform one benchmark run on the IWSLT English-Vietnamese dataset using the same parameters as in Section \ref{benchmark}. We obtain a BLEU test/dev score of 24.8/22.5 after 12K steps, reaching a score of 24.0/22.1 after 7K steps.

\subsubsection{Tensor Train}

We perform approximately 20 training runs on the IWSLT English-Vietnamese dataset with the weight matrix $W$ decomposed using Tensor Train decomposition. We use a range of maximum rank $r \in \{2,4,8,16\}$ and initial core dimensions $(m_1,m_2,m_3) \times (n_1,n_2,n_3) \in \{(2, 2, 256) \times (2, 2, 512)$, $(4, 4, 64) \times (4, 4, 128)$, $(8, 8, 16) \times (8, 8, 32)$\}. All tests were performed using the same parameters as the benchmark runs, other than the learning rate, which is specified as necessary. All runs were performed using one GPU and take approximately 1-2 hours. For the runs which obtain BLEU test $\geq$ 24.0, we report the total percentage of flops used compared with the original model. For the rest, we report the BLEU scores after 12000 training steps. The results are given in Table \ref{iwsltresults}. 

We find that the IWSLT dataset obtains a BLEU test score $\geq$ 24.0 for core dimensions $(m_1,m_2,m_3) \times (n_1,n_2,n_3) = (2, 2, 256) \times (2, 2, 512)$ with learning rate 0.0012 and rank distributions $(1,4,4,1)$ and $(1,4,16,1)$, for which we obtain BLEU test/dev = 24.0/21.9 and 24.2/21.9 respectively. These runs use 113\% and 397\% of the flops of the benchmark run respectively.  We also find in general that increasing the learning rate increases the BLEU score within a given number of steps, as does a lower $\mathrm{max}\,m_k$ and $\mathrm{max}\,n_k$. 

\begin{table}[h!]
\begin{center}
\begin{tabular}{ |c|c|c|c|c|c| } 
\hline
Rank Dist. & Weight Matrix Dimensions & Learning & BLEU & Flops \\
$(r_0,r_1,r_2,r_3)$ & $(m_1,m_2,m_3) \times (n_1,n_2,n_3)$ &  Rate & test/dev  & \%  \\
\hhline{|=|=|=|=|=|=|}
\multicolumn{3}{|l|}{\textbf{Original Model:}} & \textbf{24.0/22.4} & \textbf{100\%} \\
\hline
\multicolumn{3}{|l|}{\textbf{Low-Rank Model:}} & \textbf{24.0/22.1} & \textbf{69\%} \\
\hline
\multirow{3}{*}{$(1,2,2,1)$} & \multirow{3}{*}{$(2, 2, 256) \times (2, 2, 512)$} & 0.0012 & 23.3/21.8  & 84\% \\ & & 0.0008 & 21.7/20.1  & - \\ & & 0.0004 & 18.8/17.3 & - \\
\hline
\multirow{7}{*}{$(1,4,4,1)$}&\multirow{3}{*}{$(2, 2, 256) \times (2, 2, 512)$} & 0.0012 & \textbf{24.0/21.9} & \textbf{113\%} \\ & - & 0.0008 & 23.0/21.6 & - \\ & & 0.0004 & 21.5/19.9  & - \\
\cline{2-6}
& \multirow{3}{*}{$(4, 4, 64) \times (4, 4, 128)$} & 0.0012 & 23.0/20.8 & - \\ & & 0.0008 & 22.3/20.7 & - \\ & & 0.0004 & 20.8/19.0 & - \\ 
\cline{2-6}
& \multirow{3}{*}{$(8, 8, 16) \times (8, 8, 32)$} & 0.0012 & 22.3/20.8 & - \\ & & 0.0008 & 21.8/20.6 & - \\ & & 0.0004 & 19.9/18.6 & - \\
\hline
\multirow{3}{*}{$(1,4,8,1)$} & \multirow{3}{*}{$(2, 2, 256) \times (2, 2, 512)$} & 0.0012 & 23.9/22.1 & - \\ & & 0.0008 & 23.7/21.9 & - \\ & & 0.0004 & 23.0/21.3 & - \\
\hline
\multirow{5}{*}{$(1,8,8,1)$}& \multirow{3}{*}{$(4, 4, 64) \times (4, 4, 128)$} & 0.0012 & 23.2/21.6 & - \\ & & 0.0008 & 23.1/21.5 & - \\ & & 0.0004 & 21.8/20.4 & - \\ 
\cline{2-6}
& \multirow{3}{*}{$(8, 8, 16) \times (8, 8, 32)$} & 0.0012 & 23.1/21.1 & - \\ & & 0.0008 & 22.3/20.6 & - \\ & & 0.0004 & 20.9/19.4 & - \\
\hline
\multirow{3}{*}{$(1,4,16,1)$} & \multirow{3}{*}{$(2, 2, 256) \times (2, 2, 512)$} & 0.0012 & \textbf{24.2/21.9} & \textbf{397\%} \\ & & 0.0008 & \textbf{24.1/21.6} & \textbf{397\%} \\ & & 0.0004 & 23.3/21.5 & - \\
\hline
\multirow{2}{*}{$(1,16,16,1)$}& \multirow{3}{*}{$(4, 4, 64) \times (4, 4, 128)$} & 0.0012 & 23.4/22.0 & - \\ & & 0.0008 & 23.1/21.2 & - \\ & & 0.0004 & 22.9/21.3 & - \\
\cline{2-6}
& \multirow{2}{*}{$(8, 8, 16) \times (8, 8, 32)$}
& 0.0012 & 23.0/21.5 & - \\ & & 0.0004 & 21.3/20.0 & - \\
\hline
\end{tabular}
\caption{Training results for IWSLT English-Vietnamese '15 dataset. The weight matrix dimensions are the dimensions for the first layer (later layers have different dimensions, which are taken into account when calculating the percentage flops). For runs which obtain BLEU test $\geq$ 24.0, the total percentage of flops used compared with the original model is reported and highlighted in bold. For the rest, we report the BLEU scores after 12000 training steps.}
\label{iwsltresults}
\end{center}
\end{table}

\subsection{WMT German-English '16}

For the WMT German-English dataset, we again use hyperparameters similar to the corresponding experiment outlined in \cite{luong17}. We train 4-layer LSTMs of 1024 units with a bidirectional encoder (i.e., 2 bidirectional layers for the encoder) with embedding dimension 1024, using the Adam optimizer and a learning rate 0.0012. We train for 340K steps ($\sim$ 10 epochs) where after 170K steps, we halve the learning rate every 17K steps. The data is split into subword units using BPE (32K operations). 

We perform one training run using a rank distribution $(1,2,2,1)$ and core dimensions $(4, 4, 128) \times (4, 4, 256)$. This run was performed using 4 GPUs and took approximately 5 days. We obtain a final BLEU test/dev score of 24.0/23.8. 

We also attempted training runs using core dimensions $(2, 2, 512) \times (2, 2, 1024)$ and a maximum rank $r \in \{2,4,8\}$, also on 4 GPUs. We find that for these configurations, the application crashes due to a lack of memory. As the total number of parameters should be less than the original model when using Tensor Train decomposition, we assume this is due to intermediate copies of the matrices being stored. However, this requires further investigation.

\section{Conclusion and Future Work}\label{conclusion}

We have successfully implemented TT-layers for the TensorFlow NMT model using the t3f Tensor Train library. We have performed training runs on two datasets, the first using the IWSLT English-Vietnamese '15 dataset and the second with the WMT German-English '16 dataset. We find that the IWSLT model obtains a BLEU test score $\geq$ 24.0 for the core dimensions $(m_1,m_2,m_3) \times (n_1,n_2,n_3) = (2, 2, 256) \times (2, 2, 512)$ with learning rate 0.0012 and rank distributions $(1,4,4,1)$ and $(1,4,16,1)$, for which we obtain BLEU test/dev = 24.0/21.9 and 24.2/21.9 respectively. We also find that, of the parameters surveyed, a higher learning rate and more `rectangular' weight matrix decomposition, i.e. a lower $\mathrm{max}\,m_k$ and $\mathrm{max}\,n_k$, generally produce higher BLEU scores. We have also performed one successful training run using the larger WMT German-English dataset, using core dimensions $(4, 4, 128) \times (4, 4, 256)$ and rank distribution $(1,2,2,1)$. We obtained a final BLEU test/dev score of 24.0/23.8. 

This work shows that TT-layers can be straightforwardly introduced to the TensorFlow NMT model and can obtain BLEU scores compatible with the original. With optimization, there is potential for this model to enable more efficient model training using fewer flops, less memory and less overall training time. Training on larger datasets is currently limited by the memory consumption of the model, despite the fact that Tensor Train decomposition should use fewer parameters. This suggests that the t3f library stores intermediate copies of the matrices, which could be addressed in future work on optimization. Finally, there is further scope to apply this decomposition to the Transformer model \cite{transformer}, which produces the best BLEU results at the time of writing, with the potential to improve its efficiency.

\section*{Acknowledgements}
This work was undertaken on the Fawcett supercomputer at the Department of Applied Mathematics and Theoretical Physics (DAMTP), University of Cambridge, funded by STFC Consolidated Grant ST/P000673/1. We would like to thank Yang You for his work on the low-rank matrix format Singular Value Decomposition, and Cole Hawkins for useful comments.

AD is supported by an EPSRC iCASE Studentship  in partnership with Intel (EP/N509620/1, Voucher 16000206).

\bibliography{Paper}

\section*{Appendix}
\begin{lstlisting}[language=Python, showstringspaces=false,mathescape]
class BasicLSTMCellTT(tf.nn.rnn_cell.BasicLSTMCell):
    """ c.f. tf.nn.rnn_cell.BasicLSTMCell                                                                             
    https://github.com/tensorflow/tensorflow/blob/master/
    tensorflow/python/ops/rnn_cell_impl.py """
    def __init__(self, num_units, forget_bias=1.0,
                 state_is_tuple=True, activation=None, 
                 reuse=None, name=None):
        super(BasicLSTMCellTT, self).__init__(
            num_units, forget_bias, state_is_tuple, activation, 
            reuse, name)

    def build(self, inputs_shape):
        if inputs_shape[1].value is None:
            raise ValueError(
                "Expected inputs.shape[$-$1] to be known,
                saw shape: %s"
                % inputs_shape)

        input_depth = inputs_shape[1].value
        h_depth = self._num_units

        self._kernel = self.add_variable(
            'kernel',
            shape=[input_depth + h_depth, 4 * self._num_units])

        input_parameter = int(self._kernel.shape[0] // 4)

        kernel_TT = t3f.to_tt_matrix(self._kernel,
            shape=((2,2,input_parameter),(2,2,self._num_units)),
            max_tt_rank=4)
            
        self._kernel_TT_0 = self.add_variable(
            'kernel_TT_0',
            shape=kernel_TT.tt_cores[0].get_shape())

        self._kernel_TT_1 = self.add_variable(
            'kernel_TT_1',
            shape=kernel_TT.tt_cores[1].get_shape())

        self._kernel_TT_2 = self.add_variable(
            'kernel_TT_2',
            shape=kernel_TT.tt_cores[2].get_shape())

        self._bias = self.add_variable( 
            'bias',
            shape=[4 * self._num_units],
            initializer=tf.zeros_initializer(dtype=self.dtype))
                    
        self.built = True
        
    def call(self, inputs, state):
        sigmoid = tf.sigmoid
        one = tf.constant(1, dtype=tf.int32)
        # Parameters of gates are concatenated into one 
        # multiply for efficiency.                                                                                                      
        if self._state_is_tuple:
            c, h = state
        else:
            c, h = tf.split(value=state, num_or_size_splits=2, 
                axis=one)

        cores_tuple = (self._kernel_TT_0, self._kernel_TT_1, 
            self._kernel_TT_2)
        reconstructed_kernel_TT = t3f.TensorTrain(cores_tuple)

        gate_inputs = t3f.matmul(tf.concat([inputs, h], 1), 
            reconstructed_kernel_TT)

        gate_inputs = tf.nn.bias_add(gate_inputs, self._bias)
        # i = input_gate, j = new_input, f = forget_gate, 
        #    o = output_gate                                                                                                             
        i, j, f, o = tf.split(
            value=gate_inputs, num_or_size_splits=4, axis=one)

        forget_bias_tensor = tf.constant(self._forget_bias, 
            dtype=f.dtype)
            
        add = tf.add
        multiply = tf.multiply
        new_c = add(multiply(c, sigmoid(add(f, forget_bias_tensor))),
                    multiply(sigmoid(i), self._activation(j)))
        new_h = multiply(self._activation(new_c), sigmoid(o))

        if self._state_is_tuple:
            new_state = tf.nn.rnn_cell.LSTMStateTuple(new_c, new_h)
        else:
            new_state = tf.concat([new_c, new_h], 1)
        return new_h, new_state
        
\end{lstlisting}

\end{document}